\PassOptionsToPackage{table}{xcolor}
\PassOptionsToPackage{hyphens}{url}
\documentclass[10pt,a4paper]{article}
\usepackage{zju-ant-preprint}
\usepackage[numbers,square,sort&compress]{natbib}
\usepackage{subcaption,multirow,array,colortbl}

\renewcommand{\bibfont}{\small}
\renewcommand{\PaperTitle}{DualCast: Dual-Image Conditioned Diffusion Transformers for Digital Human E-Commerce Video Generation}
\renewcommand{\PaperShortTitle}{DualCast: Digital Human E-Commerce Video Generation}
\renewcommand{\PaperAuthors}{Lifan Jiang\textsuperscript{1,2},\quad Yibo Zhao\textsuperscript{1,2},\quad Senlin cheng\textsuperscript{2} \\[1pt]
  Bolei Xu\textsuperscript{2},\quad XIAOBO LI\textsuperscript{2},\quad Can Ye\textsuperscript{2} \\[1pt]
  Dehai Zhao\textsuperscript{1},\quad Deng Cai\textsuperscript{1},\quad Boxi Wu\textsuperscript{1,\textdagger}}
\renewcommand{\PaperAffiliations}{\textsuperscript{1}State Key Lab of CAD\&CG, Zhejiang University\qquad\textsuperscript{2}Ant Group}
\renewcommand{\PaperContact}{lifanjiang@zju.edu.cn}
\renewcommand{\PaperAuthorNote}{\textsuperscript{\textdagger}Corresponding author: Boxi Wu.}
\renewcommand{\PaperKeywords}{Digital human video generation; E-commerce; Dual-image conditioning; Diffusion Transformers}

\renewcommand{\PaperDate}{November 2026}
\hypersetup{pdfauthor={Lifan Jiang, Yibo Zhao, Senlin cheng, Bolei Xu, XIAOBO LI, Can Ye, Dehai Zhao, Deng Cai, Boxi Wu}}
\renewcommand{\PaperAbstract}{%
Digital human e-commerce video generation requires preserving product appearance, human identity, and overall video style during dynamic demonstrations. Although recent models support multiple reference images, their general conditioning schemes do not explicitly distinguish the asymmetric roles of ``what is presented'' and ``who presents it.'' Consequently, generated videos may favor a single reference, resulting in product distortion, identity drift, or single-reference collapse. To address this issue, we propose \textbf{DualCast}, a role-addressed dual-image conditioning framework for video generation. By explicitly modeling the distinct roles of product and human conditions, DualCast reduces role ambiguity and feature competition between references. First, a dual-spatiotemporal slot injection mechanism VAE-encodes the product and human images separately and places their latent features at the front and rear boundaries of the DiT sequence, enabling sequence-level condition separation and directed injection. Second, a mask-guided region-weighted loss strengthens supervision over product, face, and product--face overlap regions, improving generation fidelity in critical areas. Third, image condition dropout enables a single model to support dual-image, product-only, and human-only generation. In addition, we construct a curated dataset of 10K e-commerce videos for model training. Experimental results demonstrate the strong performance of DualCast.
}

\begin{document}
\MakeJointTitle
\begin{figure*}[t]
\centering
\includegraphics[width=0.92\textwidth]{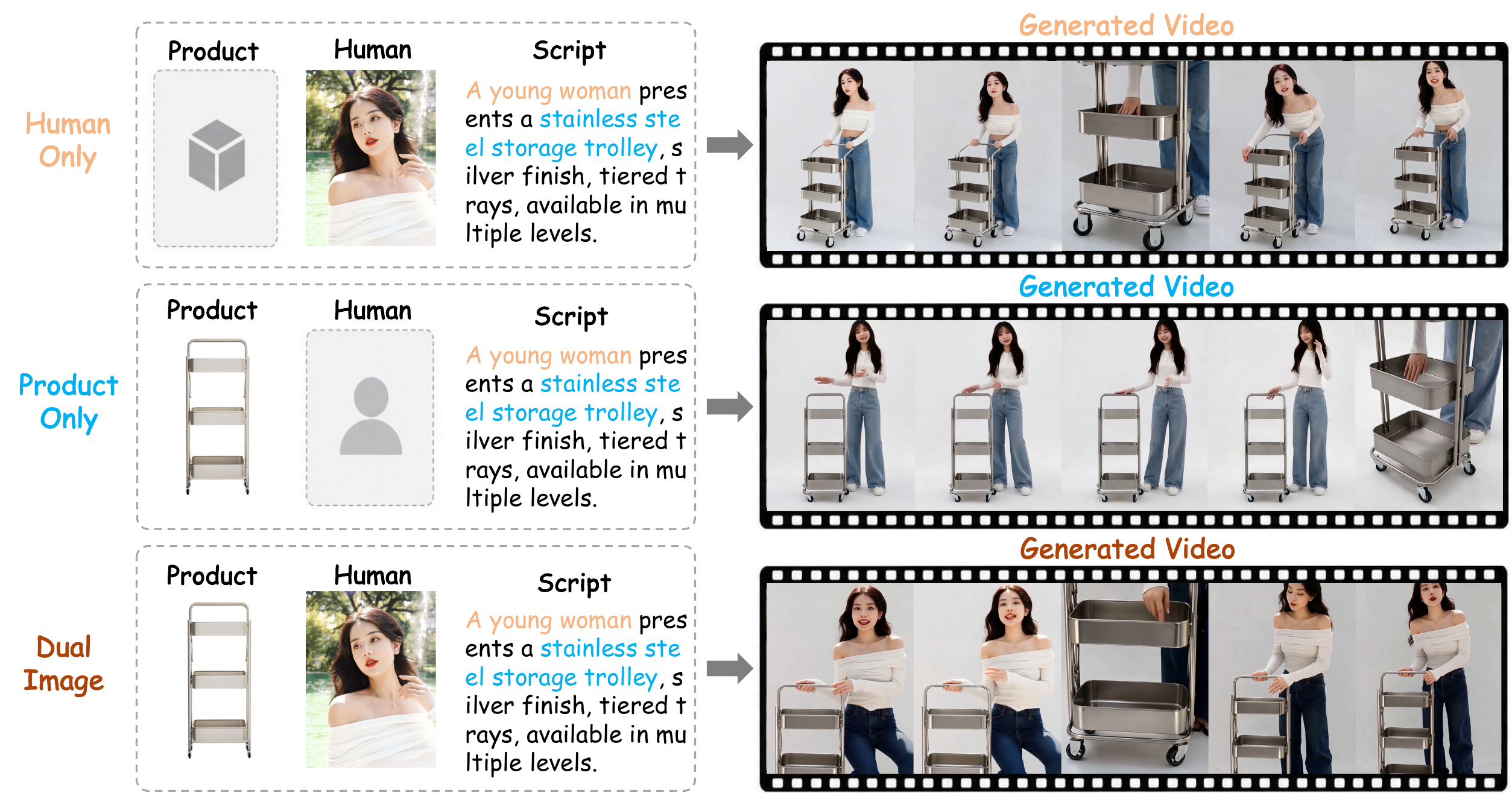}
    
    \caption{\textbf{A single fine-tuned DualCast model unifies three inference modes.} \textbf{A (Dual-image):} Using dual-spatiotemporal slot injection, the model assigns roles to and preserves both product appearance and presenter identity. \textbf{B (Product-only):} When the human condition is omitted, the model generates high-quality presenter-led promotional videos based on the product. \textbf{C (Human-only):} When the product condition is omitted, the model generates high-quality presenter-led promotional videos based on the person.}
    \label{fig:teaser}
\end{figure*}

\section{Introduction}

Digital human e-commerce video generation is conditioned on specific human and product references. Given human and product images and a text script, the goal is to generate a coherent promotional video in which a digital human introduces, displays, or demonstrates the specified product. Unlike generic text-to-video (T2V) generation, this task requires semantic and temporal coherence alongside consistent product appearance, human identity, and overall video style. Fine-grained product attributes, including logos, packaging geometry, textures, and colors, must be faithfully reproduced, while human identity should remain stable across motion and viewpoint changes. Product deformation, identity drift, or rendering errors in product--face overlap regions can significantly reduce the usability of generated videos.

Recent diffusion transformers (DiTs) and large-scale video generation models~\cite{peebles2023dit,wan2025} have demonstrated strong temporal modeling capabilities. Multi-image conditioning further enables joint control of human and product content~\cite{jiang2025vace,fei2025skyreels,zhang2025kaleido}. However, existing methods commonly inject multiple references through shared concatenation or global attention, treating them as semantically symmetric conditions. Such designs do not explicitly distinguish the asymmetric roles of ``what is presented'' and ``who presents it,'' forcing the model to infer each reference's target role from a mixed condition space. The resulting feature competition and attention bias may cause one reference to dominate, leading to product distortion, human identity drift, or single-reference collapse. This ambiguity intensifies when product and human features affect the same local regions.

To address these issues, we propose \textbf{DualCast}, a role-addressed dual-image conditioning framework for video generation. Its core idea assigns fixed, distinguishable positions to product and human conditions in the DiT sequence, establishing explicit correspondence between references and roles. Specifically, a dual-spatiotemporal slot injection mechanism independently VAE-encodes the product and human images and places their latent features at the front and rear boundaries of the DiT sequence, respectively. These boundary positions provide positional priors for both roles, enabling sequence-level condition separation and directed injection while reducing interference in shared attention. The boundary slots serve only as conditions; training loss and decoding are restricted to the intermediate video latents, leaving the output video length unchanged.

In addition to condition injection, a uniform training objective may provide insufficient supervision for local regions critical to visual fidelity. Standard flow-matching or noise-prediction losses assign equal weights to all latent locations, limiting the contribution of product, facial, and product--face overlap regions that occupy only small portions of a frame. We therefore introduce a mask-guided region-weighted loss that constructs a spatial weighting map from product and face masks and strengthens supervision over product, facial, and product--face overlap regions. This objective increases the contribution of fidelity-critical features during optimization, while loss normalization controls the training scale across samples and maintains optimization stability.

Practical applications also require flexible input configurations. Users may provide both product and human images or only one reference, while training separate models for different configurations is inefficient. We thus introduce an image condition dropout strategy. During fine-tuning, the product or human condition is randomly removed with predefined probabilities, and the corresponding boundary slots and region masks are disabled. This strategy enables a single DualCast model to support dual-image, product-only, and human-only inference. Even when one reference is absent, the model can still generate a complete digital human promotional video. In addition, we construct a curated dataset of 10K e-commerce video clips with paired product--human references and precomputed region masks for model training and evaluation. Experimental results demonstrate the strong overall performance of DualCast.

Our main contributions are summarized as follows:
\begin{itemize}
    
    \item \textbf{Role-Addressed Task Formulation:} We formulate digital human e-commerce video generation as a role-addressed dual-image conditioning problem distinguishing the asymmetric roles of product and human references.

    \item \textbf{Dual-Spatiotemporal Slot Injection:} We inject independently VAE-encoded product and human latent features into the front and rear non-generative boundary slots of the DiT sequence, using fixed positional priors to explicitly identify condition roles and reduce interference among multiple features.

    \item \textbf{Region-Weighted Supervision:} We construct a mask-guided region-weighted loss from product and face masks to strengthen supervision over product, facial, and product--face overlap regions.

    \item \textbf{Flexible Inference and E-Commerce Dataset:} We unify dual-image, product-only, and human-only inference through image condition dropout and construct a curated dataset containing 10K e-commerce video clips.
\end{itemize}

\begin{figure*}[t]
  \centering
  \includegraphics[width=\textwidth]{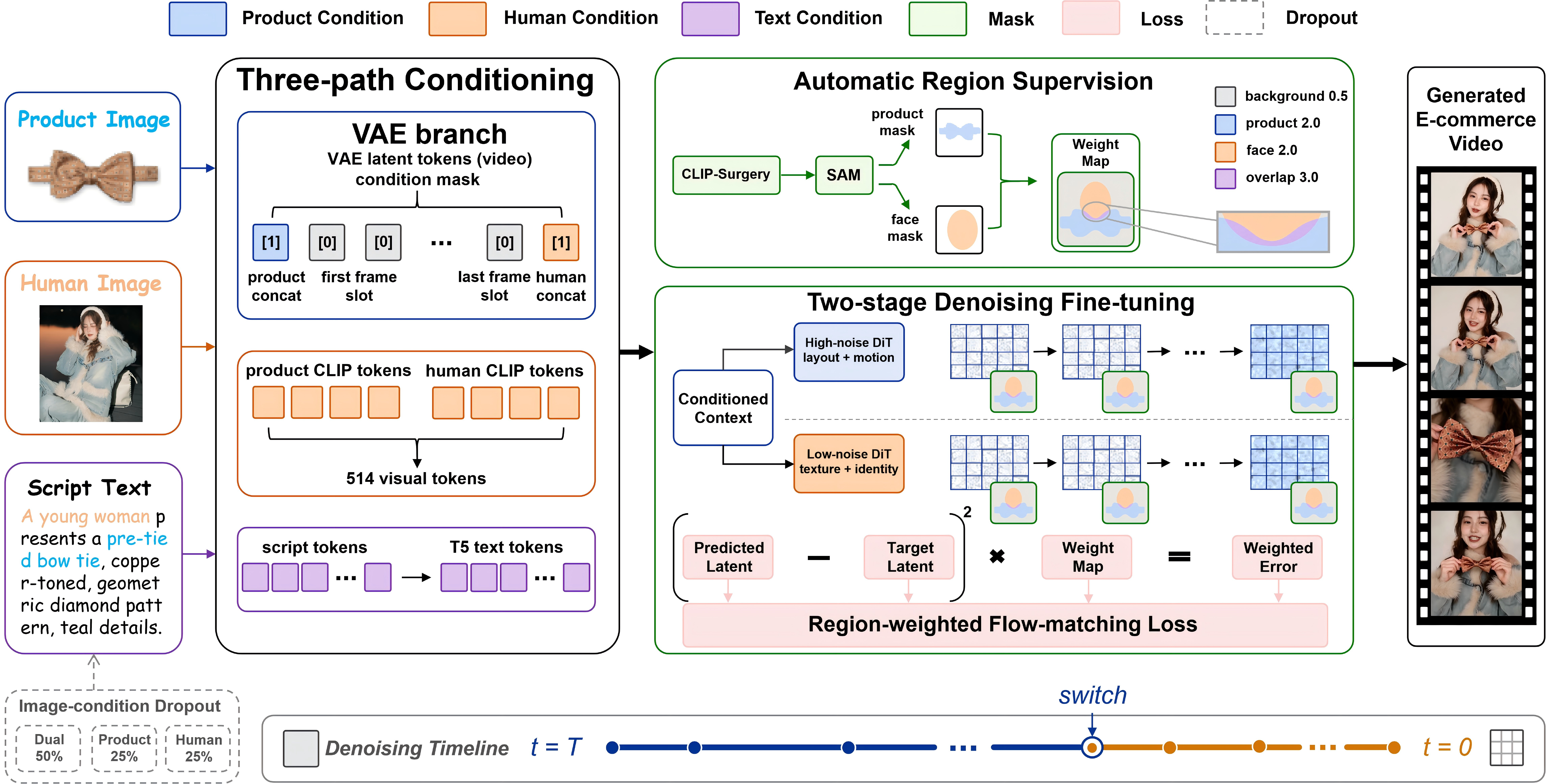}
  
    \caption{\textbf{Overview of DualCast.}
    The product and presenter images are independently VAE-encoded and placed at the front and rear boundaries of the extended latent sequence, respectively. Only intermediate video latents are denoised, supervised, and decoded. Both reference CLIP embeddings combine with the text prompt as cross-attention context. Product and face masks define region-weighted supervision, while condition dropout enables dual-image, product-only, and human-only inference.}
  \label{fig:pipeline}

\end{figure*}

\section{Related Work}

\subsection{Diffusion-Based Video Generation.}

Diffusion models~\cite{ho2020ddpm,rombach2022ldm,ho2022video} and flow-matching formulations~\cite{lipman2023flow} have transformed generative modeling, while Diffusion Transformers (DiTs)~\cite{peebles2023dit} have enabled scalable spatiotemporal synthesis. Foundation models such as Wan~\cite{wan2025} and commercial systems including Seedance 2.0, Kling-Omni, and HappyHorse~\cite{seedance2026,klingomni2025,qu2026happyhorse} have evolved beyond generic text-to-video tools into multimodal platforms supporting multi-image conditioning. Despite strong temporal priors, applying these systems to digital human e-commerce video generation remains challenging. Since references are fused as shared context, these models struggle to maintain fidelity to multiple entities over time, including branded product geometry and the stable identity of a digital human.

\subsection{Multi-Subject Reference-Driven Synthesis.}

Reference-driven methods such as ControlNet~\cite{zhang2023controlnet}, IP-Adapter~\cite{ye2023ipadapter}, and DreamBooth~\cite{ruiz2023dreambooth} have been extended to video generation~\cite{wu2023tuneavideo,guo2024animatediff}. Recent multi-subject frameworks, including VACE~\cite{jiang2025vace}, SkyReels-A2~\cite{fei2025skyreels}, Kaleido~\cite{zhang2025kaleido}, and Phantom~\cite{liu2025phantom}, further support multiple subjects or reference images. However, their conditioning interfaces do not assign persistent roles to references. In e-commerce generation, the model must therefore infer which reference defines the product and which defines the human from a shared conditioning context, leading to role ambiguity and feature competition. DualCast addresses this issue by assigning independently VAE-encoded product and human features to front and rear non-generative boundary slots. CLIP features provide shared cross-attention context, while the boundary slots offer explicit role-addressed positional priors.

\subsection{Human-Centric Animation and Interaction.}

Digital human generation has focused on identity-preserving motion transfer, talking-head synthesis, and pose-guided animation~\cite{siarohin2019fomm,prajwal2020wav2lip,zhang2023sadtalker,hu2023animateanyone,xu2024magicanimate,lin2025omnihuman}. Although these methods effectively animate facial expressions and body poses, they generally model humans independently, without explicitly preserving physical interactions with external, high-fidelity objects. In e-commerce videos, a digital human must hold, point to, or manipulate a specific product while maintaining its fine-grained appearance, introducing rendering challenges such as complex hand occlusion, local deformation, and texture bleeding around object boundaries. DualCast addresses this complementary requirement through mask-guided region-weighted supervision over product, facial, and product--face overlap regions during training. By increasing the optimization contribution of these fidelity-critical regions, the model better preserves product appearance and human identity throughout dynamic interactions.

\section{Method}

To address feature entanglement in dual-subject synthesis, we propose DualCast, a role-addressed framework (Fig.~\ref{fig:pipeline}). It comprises three core designs: (1) dual-spatiotemporal slot injection for role assignment (Sec.~\ref{sec:slot}); (2) a mask-guided region-weighted loss for region fidelity (Sec.~\ref{sec:loss}); and (3) joint training strategies for multi-mode deployment (Sec.~\ref{sec:training}). We also construct a large-scale e-commerce video dataset for training (Sec.~\ref{sec:dataset}).

\subsection{Dual-Spatiotemporal Slot Injection}
\label{sec:slot}

Standard image-to-video models typically use a single reference. Extending these interfaces to multiple references does not persistently distinguish their semantic roles. In digital human e-commerce generation, product and presenter conditions are asymmetric: the former specifies what is presented, whereas the latter specifies who presents it. We therefore assign both references to fixed boundary positions in the latent sequence. Let $I_p,I_h\in\mathbb{R}^{3\times H\times W}$ denote the product and presenter images, respectively, and $V$ a target video of $T$ frames. We encode $V$ as
\begin{equation}
x_0=\mathcal{E}_{\mathrm{VAE}}(V)
\in\mathbb{R}^{C_v\times T'\times h\times w},
\end{equation}
where $T'$ is the latent temporal length. At flow-matching timestep $\tau$, the noisy video latent is
\begin{equation}
x_\tau=(1-\tau)x_0+\tau\epsilon.
\label{eq:flow_interpolation}
\end{equation}
The references are independently encoded as
\begin{equation}
z_p=\mathcal{E}_{\mathrm{VAE}}(I_p),\qquad
z_h=\mathcal{E}_{\mathrm{VAE}}(I_h),
\end{equation}
where $z_p,z_h\in\mathbb{R}^{C_v\times1\times h\times w}$. We construct the extended latent sequence as
\begin{equation}
\bar{x}_{\tau}
=
z_p\oplus_t x_{\tau}\oplus_t z_h
\in
\mathbb{R}^{C_v\times(T'+2)\times h\times w},
\label{eq:extended_latent}
\end{equation}
where $\oplus_t$ denotes temporal concatenation. The product latent occupies the front boundary at $t=0$, while the presenter latent occupies the rear boundary at $t=T'+1$.

We define a slot indicator
$M_s\in\{0,1\}^{1\times(T'+2)\times h\times w}$,
with ones at both boundaries and zeros at intermediate positions. $M_s$ only marks the fixed boundary slots and remains unchanged under condition dropout; the product and presenter roles follow their fixed front--rear order. The indicator is concatenated with the extended latent along the channel dimension before patchification:
\begin{equation}
\tilde{x}_{\tau}
=
\operatorname{Concat}_{c}\!\left(\bar{x}_{\tau},M_s\right)
\in
\mathbb{R}^{(C_v+1)\times(T'+2)\times h\times w}.
\end{equation}
The boundary positions serve only as non-generative conditions. If $v_{\theta}(\tilde{x}_{\tau},C,\tau)$ denotes the DiT prediction over the extended sequence, we retain only its intermediate positions:
\begin{equation}
v_{\theta}^{\mathrm{gen}}
=
\Pi_{\mathrm{gen}}
\left[
v_{\theta}(\tilde{x}_{\tau},C,\tau)
\right],
\end{equation}
where $\Pi_{\mathrm{gen}}$ selects positions $1,\ldots,T'$. Training loss is computed only over these positions. During sampling, only the intermediate video latents are updated by the denoising scheduler, while boundary latents $z_p$ and $z_h$ remain fixed. Only the resulting intermediate latents are passed to the VAE decoder. Consequently, the reference slots add no output frames, and the generated video still contains $T$ frames.

\noindent\textbf{CLIP Semantic Context.}
The product and presenter images are independently processed by a CLIP vision encoder:
\begin{equation}
f_p=\operatorname{CLIP}(I_p),\qquad
f_h=\operatorname{CLIP}(I_h).
\end{equation}
Each embedding contains 257 tokens. After projection into the text-context dimension, the visual token sequences are concatenated with the T5 representation $C_{\mathrm{txt}}$ of the prompt:
\begin{equation}
C
=
\operatorname{Proj}(f_p)
\oplus
\operatorname{Proj}(f_h)
\oplus
C_{\mathrm{txt}}.
\label{eq:semantic_context}
\end{equation}
Unlike the VAE boundary latents, CLIP embeddings do not occupy temporal or generative slots. They provide auxiliary semantic information exclusively through DiT cross-attention. Explicit role assignment is realized by the fixed latent boundary positions, while the CLIP pathway improves semantic recognition of both references.

\subsection{Mask-Guided Region-Weighted Loss}
\label{sec:loss}

Standard flow-matching objectives uniformly penalize spatiotemporal locations, potentially under-emphasizing product and facial regions critical for e-commerce video generation. We therefore introduce a mask-guided region-weighted loss to strengthen supervision on these regions.

For each target video frame, we independently obtain a product mask $M_p$ and a face mask $M_f$ using CLIP Surgery \cite{li2023clipsurgery} and SAM \cite{kirillov2023sam}. Their intersection is defined after both masks are resized and binarized at the generated latent resolution. We construct the regional weight map as
\begin{equation}
W_{txy} =
\begin{cases}
\lambda_{\mathrm{overlap}}, &
M_p^{txy}=1 \land M_f^{txy}=1,\\
\lambda_p, &
M_p^{txy}=1 \land M_f^{txy}=0,\\
\lambda_f, &
M_p^{txy}=0 \land M_f^{txy}=1,\\
\lambda_{\mathrm{bg}}, &
M_p^{txy}=0 \land M_f^{txy}=0.
\end{cases}
\end{equation}
We set $\lambda_{\mathrm{overlap}}=3.0$,
$\lambda_p=2.0$, $\lambda_f=2.0$, and
$\lambda_{\mathrm{bg}}=0.5$. The overlap receives the highest weight because errors in these locations may simultaneously affect product appearance and facial identity.

Under Eq.~\ref{eq:flow_interpolation}, the target flow velocity is
\begin{equation}
u_\tau=\frac{\partial x_\tau}{\partial\tau}
=\epsilon-x_0.
\end{equation}
The region-weighted objective is computed only over the $T'$ generated latent positions:
\begin{equation}
\mathcal{L}_{\mathrm{RW}}
=
\mathbb{E}_{x_0,\epsilon,\tau}
\left[
\omega(\tau)
\frac{
\sum_{c,t,x,y}
W_{txy}
\left(
\left(v_{\theta}^{\mathrm{gen}}\right)^{ctxy}
-u_{\tau}^{ctxy}
\right)^2
}{
C_v\sum_{t,x,y}W_{txy}+\varepsilon
}
\right].
\end{equation}
Here, $v_{\theta}^{\mathrm{gen}}$ denotes the prediction restricted to the intermediate generative positions, $\omega(\tau)$ follows the timestep-weighting schedule of the base model, and $\varepsilon$ is a numerical stability constant. Normalizing by total regional weight stabilizes the loss scale across samples with varying product and face sizes.

\subsection{Multi-Mode Dropout and Split-Noise Training}\label{sec:training}
We introduce image condition dropout for multi-mode deployment and adopt a backbone-aligned fine-tuning schedule that preserves the noise-range specialization of the pretrained dual-expert DiT.

\begin{figure}[!t]
  \centering
  \includegraphics[width=0.98\columnwidth]{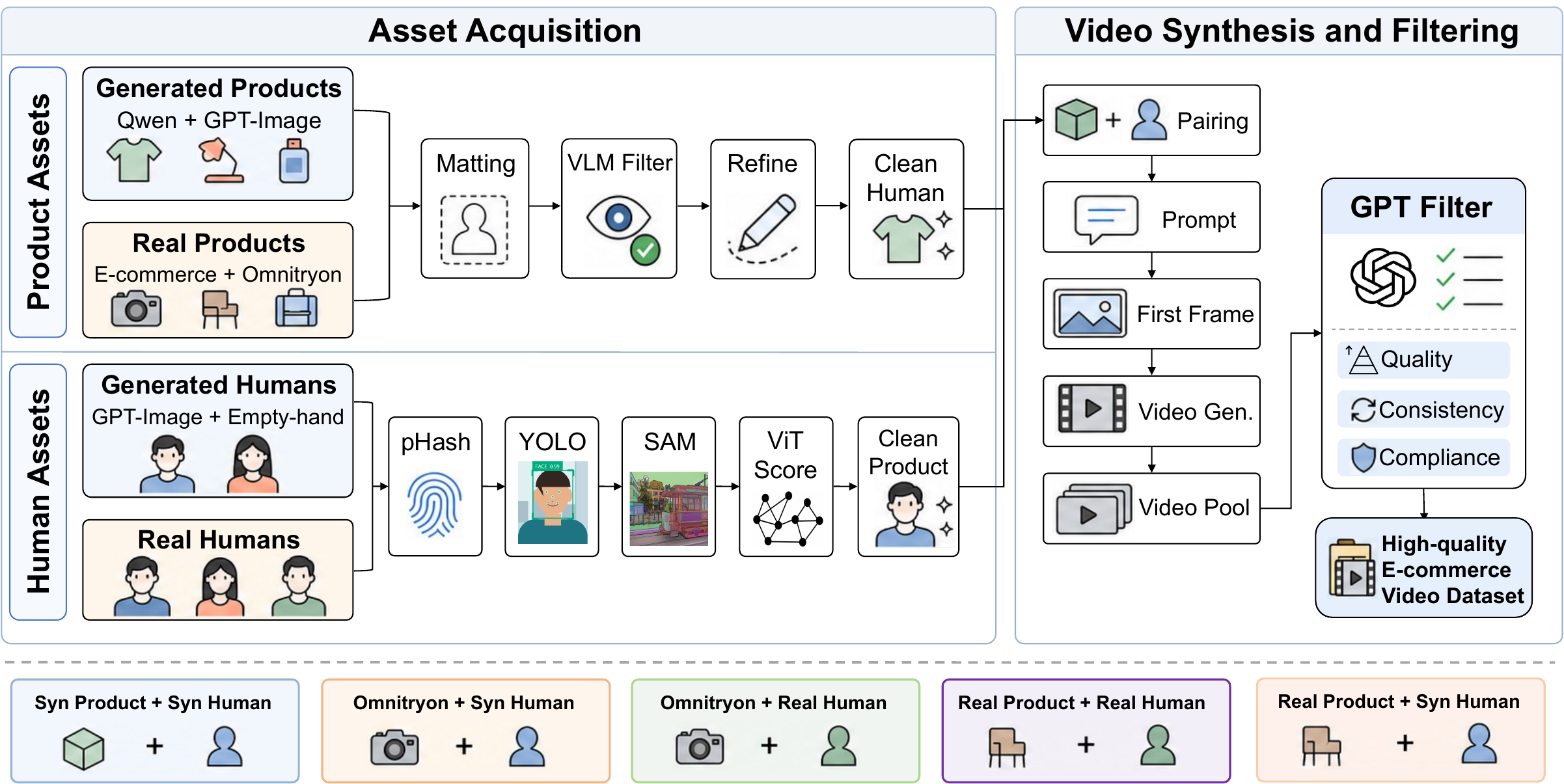}
  
    \caption{\textbf{Construction pipeline of the curated 10K e-commerce video dataset.}
    Product and presenter assets are collected or synthesized, quality-filtered, and paired to generate e-commerce-oriented prompts and initial frames. Candidate videos are subsequently filtered according to visual quality, temporal consistency, and prompt compliance. Product and face masks are precomputed for region-weighted training. Detailed source-use, filtering, privacy, and quality-control policies are provided in Supplement~\ref{app:dataset}.}
  \label{fig:data}
\end{figure}

\vspace{0.2em}
\noindent\textbf{Image Condition Dropout.}
To support three input configurations with one model, we sample dual-reference, product-only, and human-only conditions with probabilities of $50\%$, $25\%$, and $25\%$, respectively, i.e., a ratio of $2{:}1{:}1$. For dual-reference training, both boundary latents $(z_p,z_h)$ and CLIP context tokens $(f_p,f_h)$ are retained. For product-only training, $z_h$ and $f_h$ are zeroed and the face mask $M_f$ is disabled; for human-only training, $z_p$ and $f_p$ are zeroed and the product mask $M_p$ is disabled. The regional weight map $W$ is then recomputed from the remaining masks, preventing absent conditions from affecting region-weighted supervision. This joint training exposes a single checkpoint to both complete and partial visual inputs, enabling dual-image, product-only, and human-only inference.

\noindent\textbf{Backbone-Aligned Split-Noise Fine-Tuning.}
Wan2.2 employs two pretrained DiT experts specialized for different noise ranges in the denoising process. We preserve this existing specialization during fine-tuning rather than treating the two experts as a newly introduced architecture. Following Eq.~\ref{eq:flow_interpolation}, a larger $\tau$ corresponds to a higher noise level. The high-noise expert is fine-tuned only over $\tau\in[0.358,1.0]$, focusing on global composition and macroscopic motion, while the low-noise expert is fine-tuned over $\tau\in[0,0.358]$, focusing on local texture, product details, and facial appearance. Inference follows the original expert-switching boundary of the backbone.

\begin{figure*}[!t]
  \centering
  \includegraphics[width=0.96\textwidth]{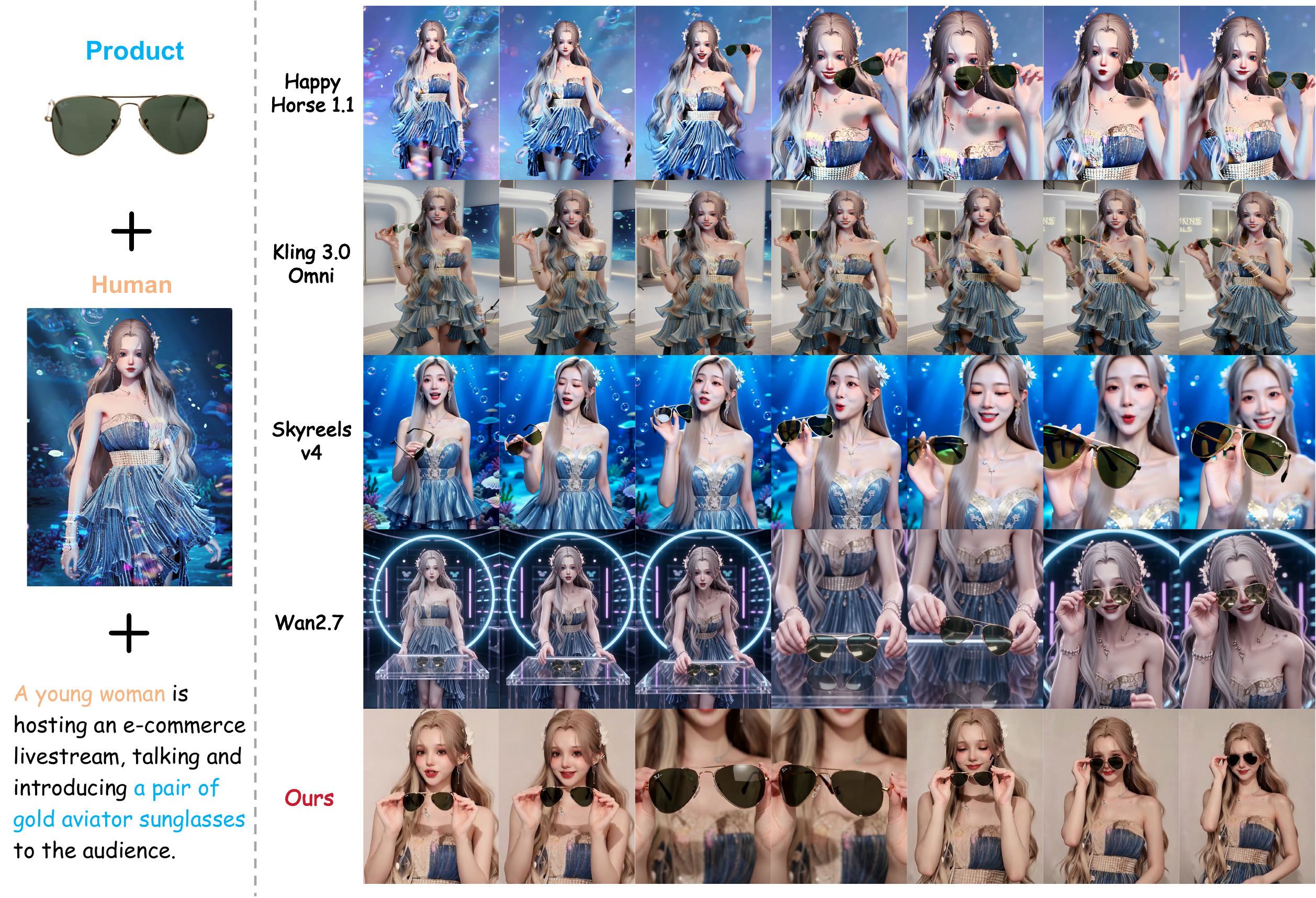}
  
    \caption{Qualitative Comparison Results. Compared with existing methods, our approach demonstrates advantages in three aspects: (1) Facial Feature Consistency --- Our method preserves the source person's facial identity during video generation, avoiding identity drift or distortion; (2) Product Consistency --- Product details from the reference image, including appearance, texture, and logos, are accurately reproduced, maintaining high visual fidelity; (3) Style Decoupling Capability --- The overall style of the generated video is not irreversibly influenced by the reference image, achieving flexible decoupling between content transfer and style control, making the generation results more suitable for e-commerce live streaming scenarios.}
  \label{fig:baseline}
\end{figure*}

\subsection{Dataset Construction}\label{sec:dataset}

We construct a 10,000-sample e-commerce dataset via two stages: \textit{Asset Acquisition} and \textit{Video Synthesis and Filtering}.

\vspace{0.1em}
\noindent\textbf{Asset Acquisition.} We derive product and human assets from both generative and real-world sources for dataset construction. For product assets, Qwen3.6-27B~\cite{yang2025qwen3technicalreport} prompts drive GPT-Image-2 to synthesize 1,500 images. Real-world products are extracted from e-commerce platforms and the Omnitryon dataset. Raw images undergo BiRefNet-based~\cite{zheng2024bilateral} foreground matting and VLM-based filtering across 8 quality dimensions. Candidates are refined via manual mask and boundary correction, yielding 1,500 real product assets combined with all clean Omnitryon products. For human assets, GPT-Image-2 synthesizes 1,500 portraits (empty-hand constraint). Real human portraits are filtered via a coarse-to-fine pipeline: pHash deduplication, YOLOv8 (single-subject constraint), SAM~\cite{kirillov2023sam} (background assessment), and a SCUT-FBP5500-finetuned ViT (aesthetic scoring)~\cite{liang2018scut}, resulting in 1,500 real human assets.

\vspace{0.1em}
\noindent\textbf{Video Synthesis and Filtering.} We execute a prompt-guided synthesis pipeline. For each product-human pair, Qwen3.6-27B generates text prompts for spatial layout and temporal dynamics. Conditioned on these, GPT-Image-2 synthesizes an initial frame, and Wan2.7 generates the video. We scale this to a large video pool by cross-pairing assets, then apply GPT-5.5 to filter samples based on visual quality, temporal consistency, and prompt compliance. We retain 10,000 high-quality samples distributed as follows: (1) 3,000 synthetic samples (1,500 original pairs + 1,500 shuffled combinations); (2) 800 Omnitryon products with synthetic humans; (3) 2,000 Omnitryon products with real humans; (4) 3,000 real e-commerce products with real humans; and (5) 1,200 real e-commerce products with synthetic humans.

\begin{table*}[!t]
\centering
\caption{\textbf{Quantitative comparison against state-of-the-art methods.}
Metrics span semantic consistency, five-dimensional perceptual quality, and video quality.
Perceptual scores are divided by 5 to obtain the 0--1 scale; 5D Avg. is their arithmetic mean.
Avg Rank$\downarrow$ averages ranks over the 11 individual metrics, excluding 5D Avg.,
with ties assigned their average ranks. All other metrics are higher-is-better.
$^\dagger$ denotes our base model. \textbf{Bold} indicates the best result;
\colorbox{green!20}{highlighted} cells denote our top-ranked metrics.}
\begingroup
\setlength{\tabcolsep}{3pt}
\resizebox{\linewidth}{!}{%
\begin{tabular}{l c | c c c | c c c c c c | c c c | c}
\toprule
\multirow{2}{*}{Method} & \multirow{2}{*}{Open}
& \multicolumn{3}{c|}{Semantic \& Consistency}
& \multicolumn{6}{c|}{Perceptual Quality (0--1)}
& \multicolumn{3}{c|}{Video Quality (VBench)}
& \multirow{2}{*}{Avg Rank$\downarrow$} \\
\cmidrule(lr){3-5}\cmidrule(lr){6-11}\cmidrule(lr){12-14}
& & ViCLIP & S2V-C & ID-Adj & EcomStyle & FaceID & ProdID & Interaction & Motion & 5D Avg. & Dynamic Deg. & Aesthetic & Imaging & \\
\midrule
Kaleido \cite{zhang2025kaleido} & \checkmark & 0.220 & 0.516 & 0.4491 & 0.904 & 0.908 & 0.828 & 0.832 & 0.860 & 0.8664 & 0.500 & 0.499 & 0.726 & 8.14 \\
Phantom \cite{liu2025phantom} & \checkmark & 0.227 & 0.493 & 0.6283 & 0.956 & 0.904 & 0.852 & 0.864 & 0.872 & 0.8896 & 0.650 & 0.510 & 0.751 & 6.95 \\
VACE \cite{jiang2025vace} & \checkmark & 0.210 & 0.596 & 0.5406 & 0.936 & 0.968 & 0.936 & 0.876 & 0.896 & 0.9224 & 0.750 & 0.513 & 0.755 & 5.86 \\
SkyReels-A2 \cite{fei2025skyreels} & \checkmark & 0.197 & 0.575 & 0.6409 & 0.876 & 0.992 & 0.860 & 0.808 & 0.768 & 0.8608 & 0.787 & 0.490 & 0.740 & 7.50 \\
\midrule
\rowcolor{blue!5}
Wan2.2-I2V-14B$^\dagger$ \textit{(Base)} \cite{wan2025} & \checkmark & 0.197 & \textbf{0.717} & 0.8206 & 0.824 & 0.796 & 0.732 & 0.716 & 0.748 & 0.7632 & 0.775 & 0.512 & 0.709 & 7.86 \\
\rowcolor{gray!15}
\textbf{DualCast (Ours)} & \checkmark & \cellcolor{green!20}\textbf{0.237} & 0.596 & \cellcolor{green!20}\textbf{0.8517} & \cellcolor{green!20}\textbf{1.000} & \cellcolor{green!20}\textbf{1.000} & \cellcolor{green!20}\textbf{0.988} & \cellcolor{green!20}\textbf{0.980} & \cellcolor{green!20}\textbf{0.956} & \cellcolor{green!20}\textbf{0.9848} & \cellcolor{green!20}\textbf{0.988} & \cellcolor{green!20}\textbf{0.524} & 0.751 & \cellcolor{green!20}\textbf{1.73} \\
\midrule
Wan2.7 \cite{aliyun_wan27_r2v} & \texttimes & 0.219 & 0.563 & 0.8498 & 0.996 & \textbf{1.000} & 0.952 & 0.960 & 0.952 & 0.9720 & 0.963 & \textbf{0.524} & 0.759 & 3.41 \\
SkyReels-v4 \cite{skyreels-v4} & \texttimes & 0.220 & 0.573 & 0.3630 & 0.988 & 0.926 & 0.816 & 0.964 & 0.924 & 0.9236 & 0.975 & 0.507 & 0.735 & 6.09 \\
HappyHorse 1.1 \cite{aliyun_happyhorse11_r2v} & \texttimes & 0.216 & 0.593 & 0.8062 & \textbf{1.000} & 0.996 & 0.984 & 0.960 & 0.932 & 0.9744 & 0.963 & 0.514 & 0.753 & 3.86 \\
Kling3.0-Omni \cite{klingomni2025} & \texttimes & 0.215 & 0.581 & 0.7744 & 0.984 & \textbf{1.000} & 0.944 & 0.976 & 0.936 & 0.9680 & 0.975 & 0.518 & \textbf{0.767} & 3.59 \\
\bottomrule
\end{tabular}
}
\endgroup
\label{tab:comparison_opensource_no_underline}
\end{table*}

\section{Experiments}

\subsection{Experimental Setup}

\noindent\textbf{Evaluation Set and Implementation.}
We construct separate 150-case sets for the main comparison and ablation studies by independently pairing product and presenter images from a held-out asset pool. All evaluation products and presenter identities are excluded from training. For each pair, Qwen3.6-27B generates an e-commerce prompt describing presentation content and human--product interaction. All methods use the same references and prompts. Videos contain 81 frames at $720\times560$; unsupported methods use the closest compatible aspect ratio. DualCast is initialized from Wan2.2-I2V-14B~\cite{wan2025}. The high- and low-noise stages are trained separately for 8 hours each on 128 NVIDIA A100 GPUs using $720\times560\times81$ videos. Dual-reference, product-only, and human-only conditions are sampled at a $2{:}1{:}1$ ratio. Region-loss weights are $0.5$, $2.0$, $2.0$, and $3.0$ for background, face, product, and product--face overlap, respectively.

\noindent\textbf{Baselines.}
We compare four open-source multi-reference methods: Kaleido~\cite{zhang2025kaleido}, Phantom~\cite{liu2025phantom}, VACE~\cite{jiang2025vace}, and SkyReels-A2~\cite{fei2025skyreels}. We also evaluate four closed-source systems: Wan2.7~\cite{aliyun_wan27_r2v}, SkyReels-v4~\cite{skyreels-v4}, HappyHorse 1.1~\cite{aliyun_happyhorse11_r2v}, and Kling3.0-Omni~\cite{klingomni2025}. All methods receive identical product and human references and role-specific prompts under their supported input protocols.

\noindent\textbf{Evaluation Metrics.}
We evaluate semantic and reference consistency, task-specific perceptual quality, and general video quality. ViCLIP measures prompt--video alignment, while S2V-C measures reference-to-video subject consistency using Grounding DINO~\cite{liu2023groundingdino}, SAM~\cite{kirillov2023sam}, and CLIP features. We additionally report ID-Adj in the main comparison, which measures facial identity consistency; higher values indicate stronger consistency. Qwen3.5-397B-A17B evaluates EcomStyle, FaceID, ProdID, Interaction, and Motion, corresponding to e-commerce style, presenter identity, product identity, product--presenter interaction, and motion naturalness. Raw 0--5 scores are normalized to $[0,1]$, and 5D Avg. is the arithmetic mean of these five normalized scores; details are provided in Supplement~\ref{app:llm_judge}. For ablations, we report ID-Ret for face-identity similarity and the additional face-detection metric FaceDet; FaceID$_{\mathrm{VLM}}$ denotes the VLM face-identity score in Table~\ref{tab:ablation_strategy}. Dynamic Degree, Aesthetic Quality, and Imaging Quality follow VBench~\cite{huang2024vbench} and DOVER~\cite{wu2023dover}.

\begin{figure}[t]
  \centering
    \includegraphics[width=0.95\linewidth]{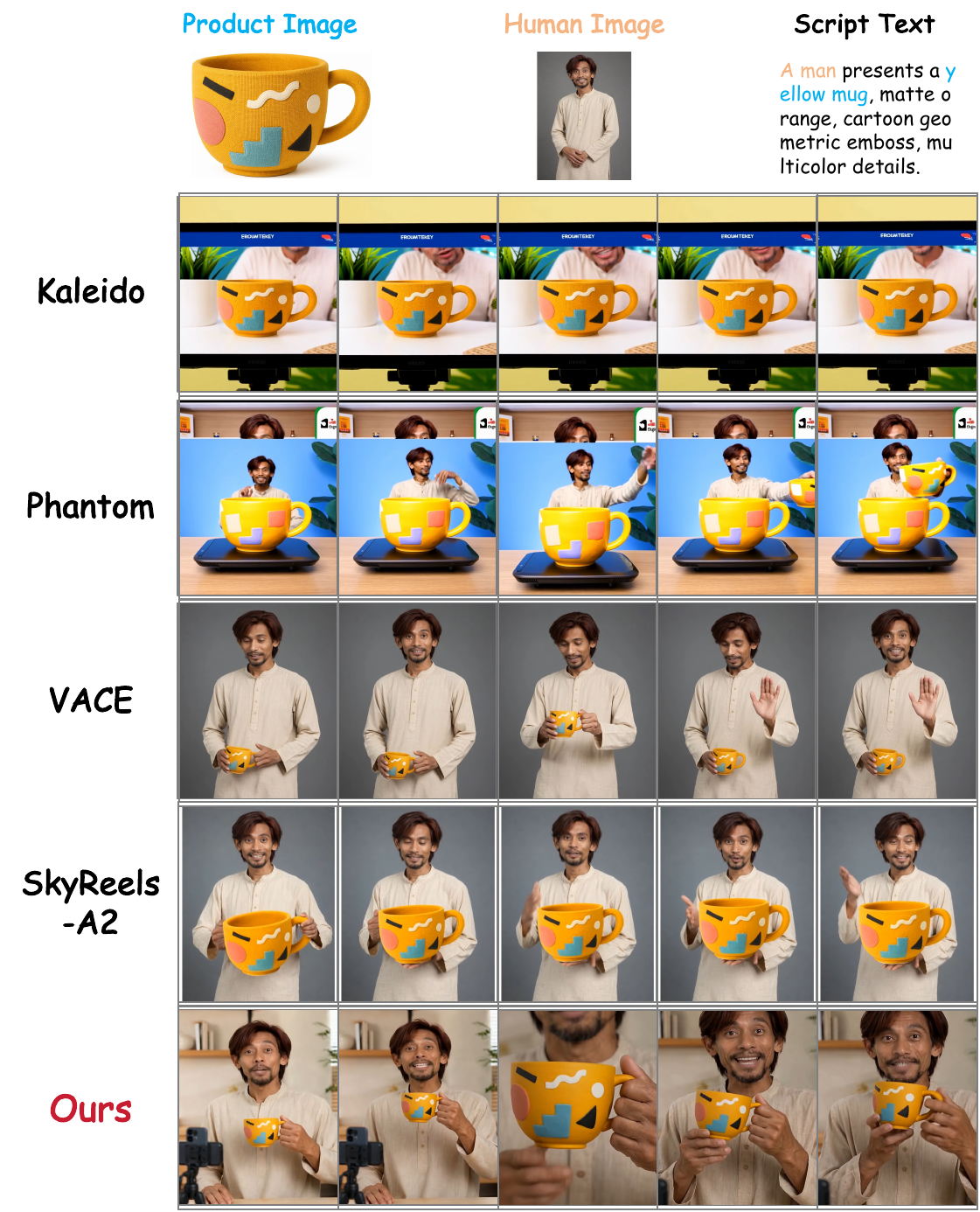}
    
    \caption{\textbf{Temporal comparison under dual-image conditioning.}
    Some baselines produce product-dominant close-ups or single-subject scenes, potentially inflating S2V-C and ProdID without preserving both references. DualCast maintains consistent product appearance and presenter identity throughout the video.}
  \label{fig:temporal_separation}
\end{figure}

\subsection{Comparison with State-of-the-Art Methods}

Table~\ref{tab:comparison_opensource_no_underline} shows that DualCast
achieves the best average rank of 1.73, ranking first or joint first
on nine of 11 individual metrics. Its ID-Adj of 0.8517 exceeds Wan2.7
(0.8498) and the base model (0.8206). The five-dimensional average
is 0.9848 (4.9240 on the original scale), ahead of HappyHorse 1.1
(0.9744) and Wan2.7 (0.9720). DualCast outperforms all closed-source
systems on ViCLIP, S2V-C, ID-Adj, ProdID, Interaction, Motion, and
Dynamic Degree. Together with Figs.~\ref{fig:baseline}
and~\ref{fig:temporal_separation}, these results demonstrate strong
reference consistency, product--presenter interaction, and video quality during dynamic product demonstrations.

\begin{table}[!t]
\centering
\caption{
  \textbf{Ablation study of the proposed components in DualCast.}
  We evaluate slot arrangements, region-loss combinations, and two-stage
  training. ProdID is divided by 5; Avg. averages all six $[0,1]$ metrics.
  \textbf{Bold} marks the best scores. Gain of Full gives relative
  changes computed before rounding; \colorbox{green!15}{green} marks positive gains.
}
\label{tab:ablation}
\begingroup
\setlength{\tabcolsep}{4pt}
\resizebox{\columnwidth}{!}{%
\begin{tabular}{l|ccc|cc|c|c}
\toprule
\multirow{2}{*}{\textbf{Model Variant}}
  & \multicolumn{3}{c|}{\textbf{Condition Fidelity}}
  & \multicolumn{2}{c|}{\textbf{Visual Quality}}
  & \multirow{2}{*}{\textbf{FaceDet}$\uparrow$}
  & \multirow{2}{*}{\textbf{Avg.}$\uparrow$} \\
\cmidrule(lr){2-4}\cmidrule(lr){5-6}
  & \textbf{ID-Ret}$\uparrow$ & \textbf{S2V-C}$\uparrow$
  & \textbf{ProdID}$\uparrow$ & \textbf{Aesthetic}$\uparrow$
  & \textbf{Imaging}$\uparrow$ & & \\
\midrule
\textbf{DualCast (Full)} & \textbf{0.5109} & \textbf{0.5886} & 0.868 & 0.52 & \textbf{0.77} & 0.7566 & \textbf{0.669} \\
\midrule
\textit{w/o} Dual-End & 0.0703 & 0.5083 & 0.856 & 0.52 & 0.74 & 0.7297 & 0.571 \\
\rowcolor{gray!12}
\textbf{Gain of Full} & \cellcolor{green!20}\textbf{$+$626.7\%} & \cellcolor{green!20}\textbf{$+$15.8\%} & \cellcolor{green!20}\textbf{$+$1.4\%} & {$\pm$0.0\%} & \cellcolor{green!20}\textbf{$+$4.1\%} & \cellcolor{green!20}\textbf{$+$3.7\%} & \cellcolor{green!20}\textbf{$+$17.2\%} \\
\midrule
\begin{tabular}[c]{@{}l@{}}\textit{w/o} Dual-End +\\\textit{w/o} Region-Loss\end{tabular} & 0.0827 & 0.5138 & 0.800 & \textbf{0.53} & 0.75 & \textbf{0.7734} & 0.575 \\
\rowcolor{gray!12}
\textbf{Gain of Full} & \cellcolor{green!20}\textbf{$+$517.8\%} & \cellcolor{green!20}\textbf{$+$14.6\%} & \cellcolor{green!20}\textbf{$+$8.5\%} & {$-$1.9\%} & \cellcolor{green!20}\textbf{$+$2.7\%} & {$-$2.2\%} & \cellcolor{green!20}\textbf{$+$16.4\%} \\
\midrule
\textit{w/o} Dual-front & 0.3041 & 0.5498 & \textbf{0.892} & 0.49 & 0.76 & 0.6985 & 0.616 \\
\rowcolor{gray!12}
\textbf{Gain of Full} & \cellcolor{green!20}\textbf{$+$68.0\%} & \cellcolor{green!20}\textbf{$+$7.1\%} & {$-$2.7\%} & \cellcolor{green!20}\textbf{$+$6.1\%} & \cellcolor{green!20}\textbf{$+$1.3\%} & \cellcolor{green!20}\textbf{$+$8.3\%} & \cellcolor{green!20}\textbf{$+$8.7\%} \\
\midrule
\begin{tabular}[c]{@{}l@{}}\textit{w/o} Dual-front +\\\textit{w/o} Region-Loss\end{tabular} & 0.4049 & 0.5690 & 0.848 & 0.51 & 0.76 & 0.7500 & 0.640 \\
\rowcolor{gray!12}
\textbf{Gain of Full} & \cellcolor{green!20}\textbf{$+$26.2\%} & \cellcolor{green!20}\textbf{$+$3.4\%} & \cellcolor{green!20}\textbf{$+$2.4\%} & \cellcolor{green!20}\textbf{$+$2.0\%} & \cellcolor{green!20}\textbf{$+$1.3\%} & \cellcolor{green!20}\textbf{$+$0.9\%} & \cellcolor{green!20}\textbf{$+$4.5\%} \\
\midrule
\textit{w/o} Region-Loss & 0.4468 & 0.5823 & 0.810 & 0.51 & 0.75 & 0.7250 & 0.637 \\
\rowcolor{gray!12}
\textbf{Gain of Full} & \cellcolor{green!20}\textbf{$+$14.3\%} & \cellcolor{green!20}\textbf{$+$1.1\%} & \cellcolor{green!20}\textbf{$+$7.2\%} & \cellcolor{green!20}\textbf{$+$2.0\%} & \cellcolor{green!20}\textbf{$+$2.7\%} & \cellcolor{green!20}\textbf{$+$4.4\%} & \cellcolor{green!20}\textbf{$+$5.0\%} \\
\midrule
\textit{w/o} Two-Stage & 0.3491 & 0.5702 & 0.868 & 0.50 & 0.74 & 0.7156 & 0.624 \\
\rowcolor{gray!12}
\textbf{Gain of Full} & \cellcolor{green!20}\textbf{$+$46.3\%} & \cellcolor{green!20}\textbf{$+$3.2\%} & {$\pm$0.0\%} & \cellcolor{green!20}\textbf{$+$4.0\%} & \cellcolor{green!20}\textbf{$+$4.1\%} & \cellcolor{green!20}\textbf{$+$5.7\%} & \cellcolor{green!20}\textbf{$+$7.2\%} \\
\bottomrule
\end{tabular}
}
\endgroup
\end{table}

\begin{table}[t]
\centering
\caption{\textbf{Ablation study on training and inference strategies.}
We compare single-modality and joint training under single- and
dual-reference inference. Scores above 1 are divided by 5;
values already in $[0,1]$ are retained.
FaceID$_{\mathrm{VLM}}$ denotes VLM-based face-identity assessment.
\textbf{Bold} marks the best values within each group;
\colorbox{green!15}{green} highlights gains over the corresponding
dual-inference baseline. Avg. averages the four metrics;
$^\dagger$ excludes unavailable entries.}
\label{tab:ablation_strategy}
\begingroup
\setlength{\tabcolsep}{3pt}
\resizebox{\linewidth}{!}{%
\begin{tabular}{l l | c c c c | c}
\toprule
\textbf{Training} & \textbf{Inference} & \textbf{ID-Ret}$\uparrow$
& \textbf{EcomStyle}$\uparrow$ & \textbf{FaceID}$_{\mathrm{VLM}}\!\uparrow$
& \textbf{ProdID}$\uparrow$ & \textbf{Avg.}$\uparrow$ \\
\midrule
\multicolumn{7}{c}{\textit{Product-Centric Comparison}} \\
\midrule
Baseline (Product-Only) & Product-Only & 0.0645 & 0.960 & 0.2120 & 0.838 & 0.519 \\
Baseline (Product-Only) & Dual & 0.4260 & 0.904 & 0.7660 & 0.808 & 0.726 \\
\midrule
Ours (Joint) & Product-Only & {---} & \textbf{0.980} & 0.9600 & 0.862 & \textbf{0.934}$^\dagger$ \\
Ours (Joint) & \textbf{Dual} & \textbf{0.5109} & \textbf{0.980} & \textbf{0.9840} & \textbf{0.868} & 0.836 \\
\rowcolor{gray!12}
\textbf{$\Delta$ vs. Baseline (Dual)} & ~ & \cellcolor{green!20}\textbf{$+$19.9\%} & \cellcolor{green!20}\textbf{$+$8.4\%} & \cellcolor{green!20}\textbf{$+$28.5\%} & \cellcolor{green!20}\textbf{$+$7.4\%} & \cellcolor{green!20}\textbf{$+$15.1\%} \\
\midrule
\multicolumn{7}{c}{\textit{Human-Centric Comparison}} \\
\midrule
Baseline (Human-Only) & Human-Only & 0.4682 & 0.968 & 0.9720 & 0.454 & 0.716 \\
Baseline (Human-Only) & Dual & 0.3910 & 0.940 & 0.8460 & 0.817 & 0.749 \\
\midrule
Ours (Joint) & Human-Only & 0.4983 & 0.979 & \textbf{0.9841} & {---} & 0.820$^\dagger$ \\
Ours (Joint) & \textbf{Dual} & \textbf{0.5109} & \textbf{0.980} & 0.9840 & \textbf{0.868} & \textbf{0.836} \\
\rowcolor{gray!12}
\textbf{$\Delta$ vs. Baseline (Dual)} & ~ & \cellcolor{green!20}\textbf{$+$30.7\%} & \cellcolor{green!20}\textbf{$+$4.3\%} & \cellcolor{green!20}\textbf{$+$16.3\%} & \cellcolor{green!20}\textbf{$+$6.2\%} & \cellcolor{green!20}\textbf{$+$11.7\%} \\
\bottomrule
\end{tabular}
}
\endgroup
\end{table}

\begin{table}[t]
\centering
\caption{\textbf{User-study preference rates (\%).}
Values compare DualCast with each baseline; rates above 50\% favor
DualCast, with ties counted as 0.5. $^\ast$ denotes statistical
significance at $p<0.05$. See Supplement~\ref{app:user_study} for details.}
\label{tab:user_study}

\resizebox{\columnwidth}{!}{
\begin{tabular}{l|cccc}
\toprule
\textbf{Evaluation Metric} & \textbf{vs. Kaleido} & \textbf{vs. Phantom} & \textbf{vs. VACE} & \textbf{vs. SkyReels-A2} \\
\midrule
Product Consistency & 63.6\%$^\ast$ & 67.2\%$^\ast$ & 52.0\%$^\ast$ & 67.3\%$^\ast$ \\
Identity Preservation & 76.8\%$^\ast$ & 79.8\%$^\ast$ & 60.2\%$^\ast$ & 61.6\%$^\ast$ \\
Video Quality & 71.4\%$^\ast$ & 72.2\%$^\ast$ & 67.7\%$^\ast$ & 82.4\%$^\ast$ \\
\midrule
\rowcolor{gray!10}\textbf{Overall Preference} & 70.6\% & 73.1\% & 60.0\% & 70.4\% \\
\bottomrule
\end{tabular}
}
\end{table}

\subsection{Ablation Study}

Tables~\ref{tab:ablation} and~\ref{tab:ablation_strategy} validate the effectiveness of the proposed architecture and training strategies.

\noindent\textbf{Effect of Dual-Spatiotemporal Slot Injection.}
Removing the dual-end arrangement reduces ID-Ret from 0.5109 to 0.0703
and S2V-C from 0.5886 to 0.5083. The full model improves their scores
by 626.7\% and 15.8\%, respectively. Against \textit{w/o Dual-front},
it improves ID-Ret by 68.0\%, S2V-C by 7.1\%, and Imaging
Quality by 1.3\%, with lower ProdID. It achieves the highest six-metric average.

\noindent\textbf{Effect of Region-Weighted Loss.}
With the full slot arrangement, region weighting improves ID-Ret by
14.3\%, S2V-C by 1.1\%, and ProdID by 7.2\%, increasing the average
by 5.0\%. Its effect depends on slot arrangement: removing it from
either \textit{w/o Dual-End} or \textit{w/o Dual-front} improves
ID-Ret but reduces ProdID. Thus, the combined ablations reveal
configuration-dependent tradeoffs.

\noindent\textbf{Effect of Split-Noise Fine-Tuning.}
Against \textit{w/o Two-Stage}, full training improves ID-Ret by
46.3\%, S2V-C by 3.2\%, Aesthetic Quality by 4.0\%, Imaging Quality
by 4.1\%, and FaceDet by 5.7\%, maintaining ProdID and improving
the average by 7.2\%.

\noindent\textbf{Face Detection and Identity Preservation.}
The highest FaceDet (0.7734) occurs with ID-Ret of only 0.0827,
whereas the full model has FaceDet of 0.7566 and ID-Ret of 0.5109.
Detection alone therefore does not establish identity preservation.

\noindent\textbf{Effect of Image Condition Dropout.}
Table~\ref{tab:ablation_strategy} shows dual-inference average gains
of 15.1\% and 11.7\% over product-only and human-only training,
with ID-Ret gains of 19.9\% and 30.7\% and VLM face-identity gains
of 28.5\% and 16.3\%, respectively. Against human-only training,
ProdID improves by 6.2\%. The same checkpoint supports product-only
inference (ProdID 0.862) and human-only inference (ID-Ret 0.4983) using identical model parameters.

\noindent\textbf{User Study.}
DualCast achieves overall preference rates of 60.0\%--73.1\%
(Table~\ref{tab:user_study}); protocols are in Supplement~\ref{app:user_study}.

\begin{table}[t]
\centering
\caption{\textbf{Full-video temporal diagnostics of DualCast.}
All 150 videos are covered by 1050 overlapping windows. Visual
flags are MLLM-suspected events; counts denote videos with each
flag, not manually confirmed failures. Scores use exclusive-frame
weights; motion statistics are computed per video. Details are in
Supplement~\ref{app:temporal_diagnostics}.}
\label{tab:temporal_diagnostics}
\begingroup
\small
\setlength{\tabcolsep}{3pt}
\begin{tabular*}{\columnwidth}{@{\extracolsep{\fill}}lr@{}}
\toprule
\textbf{Diagnostic} & \textbf{Result} \\
\midrule
\multicolumn{2}{l}{\textit{Execution and coverage}} \\
Video outputs & 150/150 \\
Completed diagnostic windows & 1050/1050 \\
Decoded-frame coverage & 100\% \\
API / file / decoding errors & 0 / 0 / 0 \\
\midrule
\multicolumn{2}{l}{\textit{Content and temporal checks}} \\
Identity drift & 0/150 (0.0\%) \\
Reference copying & 0/150 (0.0\%) \\
Product deformation & 14/150 (9.3\%) \\
Ghosting & 13/150 (8.7\%) \\
Interpenetration & 3/150 (2.0\%) \\
Camera shake & 1/150 (0.7\%) \\
\midrule
\multicolumn{2}{l}{\textit{Window-weighted scores (0--5)}} \\
E-commerce style (mean) & 4.925 \\
Face identity (mean) & 4.794 \\
Product identity (mean) & 4.748 \\
Human--product interaction (mean) & 4.920 \\
Motion naturalness (mean) & 4.008 \\
Motion naturalness (median) & 4.15 \\
Videos with motion score $\geq3.5$ & 83\% \\
\bottomrule
\end{tabular*}
\endgroup
\end{table}

\subsection{Full-Video Temporal Robustness}
\label{sec:temporal_robustness}

We diagnose all decoded frames of 150 DualCast videos using overlapping
16-frame windows (stride 12) and the same multimodal judge.
Table~\ref{tab:temporal_diagnostics} reports complete processing with
no API, file, or decoding errors. The judge flags no identity drift
or reference copying; product deformation and ghosting affect fewer
than 10\% of videos each, with interpenetration and camera shake
less frequent. Mean face- and product-identity scores are 4.794 and
4.748 out of 5; the median motion score is 4.15, with 83\% of videos
scoring at least 3.5. These results support reliable execution and
consistent subject representation under this diagnostic protocol. Detailed event
counts, five-axis scores, and segment-wise results are provided in
Supplement~\ref{app:temporal_diagnostics}.

\section{Conclusion}

We presented DualCast, a role-addressed dual-image conditioning framework for digital human e-commerce video generation. Motivated by the asymmetric functions of product and presenter references, DualCast augments shared semantic conditioning with a consistent role-addressed latent structure. The proposed dual-boundary slot injection separates the two visual references, the mask-guided region-weighted objective emphasizes task-critical subject regions, and image-condition dropout enables a single fine-tuned model to operate under dual-reference, product-only, and human-only configurations. Experiments and user studies demonstrate that DualCast provides a strong balance among prompt alignment, subject consistency, visual quality, and joint product--presenter fidelity, while the ablation results verify the contributions of its main components.

\noindent\textbf{Limitations and Future Work.}
The current framework assumes a single product and a single presenter with fixed role assignments, limiting its flexibility for more complex e-commerce scenarios involving multiple products, presenters, or tasks. Future work will explore more flexible role representations and extend DualCast to multi-product, multi-presenter, and long-form video generation with additional controls such as pose, speech, and camera trajectories.

\bibliographystyle{ieeenat_fullname}
\bibliography{main}
\clearpage
\appendix

\section{Full-Video Temporal Diagnostic Protocol}
\label{app:temporal_diagnostics}

\subsection{Scope and Complete Frame Coverage}

The M06 diagnostic evaluates the 150 saved DualCast outputs
(\texttt{dual\_final}). Each video contains 81 frames, giving
12,150 decoded frames in total. The diagnostic uses Qwen3.5-397B-A17B,
the judge employed for the main evaluation, on overlapping windows
of at most 16 frames with a four-frame overlap and stride 12.
The seven windows start at frames 0, 12, 24, 36, 48, 60, and 72;
the last window covers frames 72--80. Their union covers every decoded
frame, including the final frame. Each window includes both reference
images, the text script, and absolute frame numbers and timestamps.
The window union establishes complete coverage of the decoded
frames submitted to the judge.

All 150 output files are present and decode successfully. The 1050
diagnostic windows complete with zero API errors. Window-level records
store event evidence and support resumable processing.

\subsection{Event Counts and Interpretation}

Table~\ref{tab:temporal_event_details} expands the event counts in
Table~\ref{tab:temporal_diagnostics}. The MLLM assigns suspected event
labels; these labels have not been human-verified. Video counts are
deduplicated within each type, with overlap allowed across types.
Frame counts are also recorded by event type. Product-change labels
in this snapshot identify product deformation. These conventions
preserve the event-level evidence and its scope.

\begin{table}[!ht]
\centering
\caption{\textbf{Detailed visual-event diagnostics.}
Video counts are deduplicated within each MLLM event type.
The diagnostic covers 12,150 frames.}
\label{tab:temporal_event_details}
\begingroup
\small
\setlength{\tabcolsep}{3pt}
\begin{tabular*}{\columnwidth}{@{\extracolsep{\fill}}lrrr@{}}
\toprule
\textbf{Event type} & \textbf{Videos} & \textbf{Events} & \textbf{Frames} \\
\midrule
Product deformation & 14 & 15 & 114 \\
Ghosting & 13 & 13 & 62 \\
Interpenetration & 3 & 3 & 9 \\
Camera shake & 1 & 1 & 3 \\
Identity drift & 0 & 0 & 0 \\
Reference copying & 0 & 0 & 0 \\
\bottomrule
\end{tabular*}
\endgroup
\end{table}

\subsection{Window-Aggregated Scores}

The seven window scores are weighted by their exclusive frame counts, $12,12,12,12,12,12,9$,
which sum to 81. If $s_{i,k,a}$ is the score for video $i$, window $k$,
and axis $a$, its per-video score is
\begin{equation}
\bar{s}_{i,a}=\frac{1}{81}\sum_{k=1}^{7}w_k s_{i,k,a}.
\end{equation}
The reported mean averages these per-video scores over 150 videos.
Table~\ref{tab:temporal_axis_scores} reports the resulting 0--5
scores. The main comparison uses the six-frame protocol in
Supplement~\ref{app:llm_judge}.

\begin{table}[!ht]
\centering
\caption{\textbf{Five-axis full-video diagnostic scores.}
Means use exclusive-frame weighting within each video and then
average over all 150 videos.}
\label{tab:temporal_axis_scores}
\begingroup
\small
\setlength{\tabcolsep}{3pt}
\begin{tabular*}{\columnwidth}{@{\extracolsep{\fill}}lr@{}}
\toprule
\textbf{Axis} & \textbf{Mean (0--5)} \\
\midrule
E-commerce style & 4.925 \\
Face identity & 4.794 \\
Product identity & 4.748 \\
Human--product interaction & 4.920 \\
Motion naturalness & 4.008 \\
\bottomrule
\end{tabular*}
\endgroup
\end{table}

For the per-video motion scores, the median is 4.15, with P25/P75
of 3.67/4.48 and a range of 0.85--5.00. The snapshot reports 83\%
of videos scoring at least 3.5, 61\% scoring at least 4.0, and
16/150 videos (10.7\%) below 3.0.

\subsection{Scores Across Temporal Segments}

Table~\ref{tab:temporal_segment_scores} reports the first, middle,
and last 27-frame segments. Face and product identity scores remain
at least 4.597 and 4.680, respectively, across segments, while the
human--product interaction score remains at least 4.901. Motion scores
vary more, with the middle segment lower than the beginning and end.
The segment scores localize variation within each diagnostic axis.

\begin{table}[!ht]
\centering
\caption{\textbf{Segment-wise diagnostic scores (0--5).}
The first, middle, and last segments each contain 27 frames.
Inter. denotes human--product interaction.}
\label{tab:temporal_segment_scores}
\begingroup
\small
\setlength{\tabcolsep}{2pt}
\begin{tabular*}{\columnwidth}{@{\extracolsep{\fill}}lrrrrr@{}}
\toprule
\textbf{Segment} & \textbf{Style} & \textbf{Face} & \textbf{Product} & \textbf{Inter.} & \textbf{Motion} \\
\midrule
First & 4.928 & 4.876 & 4.766 & 4.923 & 4.029 \\
Middle & 4.887 & 4.597 & 4.680 & 4.901 & 3.725 \\
Last & 4.961 & 4.908 & 4.797 & 4.936 & 4.270 \\
\bottomrule
\end{tabular*}
\endgroup
\end{table}

\section[Inference Configuration and Computational Cost]{\raggedright Inference Configuration and Computational Cost}
\label{app:inference_cost}

\subsection{Hardware and Inference Configuration}

We report the fixed inference-benchmark snapshot for Wan2.2-I2V-A14B
under three input modes: dual-image (product and presenter), human-only,
and product-only. Table~\ref{tab:inference_config} lists the shared
configuration. The high- and low-noise networks are dense DiTs
specialized for their corresponding noise ranges. Both remain
resident on the GPU throughout inference, with offloading disabled.

\begin{table}[!ht]
\centering
\caption{\textbf{Shared inference-benchmark configuration.}}
\label{tab:inference_config}
\begingroup
\footnotesize
\setlength{\tabcolsep}{3pt}
\begin{tabular}{@{}>{\raggedright\arraybackslash}p{0.33\columnwidth}>{\raggedright\arraybackslash}p{0.61\columnwidth}@{}}
\toprule
\textbf{Setting} & \textbf{Value} \\
\midrule
Hardware & 4 NVIDIA A100-SXM4-80GB GPUs \\
Precision & \texttt{bfloat16} \\
Offloading & None \\
Parameter residency & \texttt{all\_params\_gpu\_resident} \\
Resolution & $560\times720$ pixels (as logged) \\
Frames / frame rate & 81 / 15 FPS \\
Random seed & 42 \\
DiT switch argument & \texttt{switch\_DiT\_boundary} $=0.9$ \\
High-noise LoRA & \texttt{epoch-7.safetensors}\newline Target: \texttt{dit} \\
Low-noise LoRA & \texttt{epoch-7.safetensors}\newline Target: \texttt{dit2} \\
Mean model load time & Approximately 41--42\,s \\
\bottomrule
\end{tabular}
\endgroup
\end{table}

\subsection{Measurement Protocol}

Each latency measurement is the actual elapsed time for one inference
run, with CUDA synchronization at the timing boundaries. Image loading,
output saving, and model loading are excluded. The model-loading time
in Table~\ref{tab:inference_config} is reported separately.

Peak memory is measured using
\path{torch.cuda.max_memory_allocated}, which records peak memory
allocated by the current process through the CUDA allocator. Values
are reported in GiB for the current process on the measured device.

\subsection{Cost Across Input Modes}

Table~\ref{tab:inference_cost} summarizes the logged runs, pooled across
six benchmarked methods within each input mode. The snapshot contains
60 cases for each of the dual-image, human-only, and product-only modes.
The reported statistics describe the workload pooled across these
methods. The quality evaluations in the main paper use 150 cases.

\begin{table}[!ht]
\centering
\caption{\textbf{Inference cost by input mode, pooled across methods.}
Latency excludes image I/O and model loading; memory denotes
process-level peak allocated memory. Percentiles use linear
interpolation.}
\label{tab:inference_cost}
\begingroup
\small
\setlength{\tabcolsep}{3pt}
\begin{tabular}{@{}lrrr@{}}
\toprule
\textbf{Statistic} & \textbf{Dual} & \textbf{Human} & \textbf{Product} \\
\midrule
Cases & 60 & 60 & 60 \\
Methods & 6 & 6 & 6 \\
\midrule
Mean latency (s) & 867.0 & 869.0 & 865.6 \\
Median latency (s) & 866.3 & 868.9 & 864.7 \\
P10 latency (s) & 865.5 & 867.8 & 863.7 \\
P25 latency (s) & 865.8 & 868.4 & 864.0 \\
P75 latency (s) & 868.8 & 869.3 & 867.6 \\
P90 latency (s) & 869.2 & 869.9 & 868.1 \\
\midrule
Mean peak memory (GiB) & 68.61 & 68.61 & 68.61 \\
Maximum peak memory (GiB) & 68.62 & 68.62 & 68.62 \\
\bottomrule
\end{tabular}
\endgroup
\end{table}

Mean latency ranges from 865.6 to 869.0\,s, with a 3.4\,s difference
between modes. The P10--P90 spans are 3.7\,s, 2.1\,s, and 4.4\,s
for dual-image, human-only, and product-only inference, respectively.
For dual-image inference, the middle 50\% of latencies lie between
865.8 and 868.8\,s. The P10--P90 intervals summarize the central 80\% of measured
latencies. Mean and maximum peak allocated memory are 68.61 and
68.62\,GiB in every mode, respectively.

\subsection{Benchmark Records}

\begingroup
\raggedright
The statistics are taken from a fixed snapshot of per-mode JSONL
records matching \path{metrics/inference_bench/e11/logs/<mode>/*.jsonl}.
They are summarized with
\path{examples/wanvideo/model_inference/bench_inference_patterns.py}
using \texttt{--summarize-only --mode all} and the benchmark's
configuration arguments. The snapshot stores the combined results
for all three modes.
\par
\endgroup

\section{User Study Protocol}
\label{app:user_study}

This section describes the human-evaluation protocol used for
Table~\ref{tab:user_study}. We conduct a within-subjects, blind comparison
between DualCast and four baselines: Kaleido, Phantom, VACE, and
SkyReels-A2.

\subsection{Participants}

Fifty evaluators participated in the study. Each evaluator registered
with a nickname that was converted into an anonymous hashed identifier.
No personally identifiable information was collected. All evaluators
received the same instructions and used an identical evaluation interface.

\subsection{Stimuli and Procedure}

The study uses a fixed pool of 50 cases selected before evaluation.
Each case contains a product reference, a presenter reference, a text
prompt, and videos generated by the five methods. All videos are
normalized to $720\times560$ resolution, 81 frames, a five-second
duration, and no audio.

Each evaluator is assigned 20 cases using a deterministic random seed
derived from the anonymous identifier. This allows an evaluator to resume
the same cases after re-entering the website. For each case, all five
videos are displayed side by side and labeled only as ``Video 1'' through
``Video 5.'' Their left-to-right order is independently randomized for
each evaluator and case to reduce positional bias.

\subsection{Rating Criteria}

Evaluators independently rate every video on a five-point ordinal scale,
where 1 denotes the worst performance and 5 denotes the best, according
to the following dimensions:
\begin{enumerate}
    \item \textbf{Product Consistency:} similarity between the generated
    product and the reference product;
    \item \textbf{Identity Preservation:} identity consistency between
    the generated presenter and the reference portrait;
    \item \textbf{Video Quality:} overall visual quality, motion
    naturalness, and the absence of visible artifacts.
\end{enumerate}

The study collects 15,000 ratings in total:
$50$ evaluators $\times$ $20$ cases $\times$ $5$ methods $\times$
$3$ dimensions.

\subsection{Aggregation and Statistical Testing}

For a given evaluator, case, dimension, and baseline, let
$s_{\mathrm{ours}}$ and $s_b$ denote the scores of DualCast and the
baseline. The pairwise outcome is defined as
\begin{equation}
w =
\begin{cases}
1,   & s_{\mathrm{ours}} > s_b,\\
0.5, & s_{\mathrm{ours}} = s_b,\\
0,   & s_{\mathrm{ours}} < s_b.
\end{cases}
\end{equation}
The preference rate is the mean of $w$ over all paired comparisons,
multiplied by $100\%$. Thus, a rate above $50\%$ favors DualCast, while
ties contribute 0.5. Each baseline and evaluation dimension contains 1,000 evaluator--case comparisons for computing preference rates. For significance testing, we first average the paired score differences over the 20 cases assigned to each evaluator, producing 50 evaluator-level differences for each baseline and dimension. Statistical significance is then assessed using a two-sided Wilcoxon signed-rank test on these evaluator-level differences. In Table~\ref{tab:user_study},
$^\ast$ denotes $p<0.05$.

\section{Multimodal LLM-as-a-Judge Protocol}
\label{app:llm_judge}

\subsection{Overview}

Qwen3.5-397B-A17B assesses e-commerce presentation style, subject
identity, human--product interaction, and motion naturalness. Given sampled video frames and
the corresponding human and product references, the judge independently
scores five dimensions on an integer scale from 0 to 5, where higher is
better.

\subsection{Evaluation Dimensions}

\noindent\textbf{EcomStyle.}
This metric evaluates whether the video follows the visual and
presentational conventions of e-commerce broadcasting, including a
camera-facing presenter, clear product presentation, and appropriate
composition. A score of 5 indicates a convincing e-commerce presentation,
3 indicates weak selling intent, and 0 indicates an unrelated or invalid
scene.

\noindent\textbf{FaceID.}
This metric evaluates whether the generated presenter has the same facial
identity as the human reference. The judge focuses on facial structure and
proportions while ignoring clothing, hairstyle, accessories, background,
pose, and minor illumination changes. Scores of 5, 3, and 0 indicate the
same identity, an uncertain identity, and a different identity,
respectively.

\noindent\textbf{ProdID.}
This metric evaluates whether the generated product matches the reference
in category, shape, color, texture, and branding or packaging, while
allowing viewpoint and scale changes. Scores of 5, 3, and 0 indicate the
same product, an uncertain match, and a different or missing product,
respectively.

\noindent\textbf{Interaction.}
This metric evaluates whether the presenter actively holds, points to,
displays, or explains the target product. A score of 5 indicates a natural
selling interaction, 3 indicates a weak association, and 0 indicates that
the human and product are unrelated.

\noindent\textbf{Motion.}
This metric evaluates lip movement, gesture variation, pose changes,
and temporal artifacts such as jitter and ghosting. Scores of 5, 3,
and 0 indicate natural motion, limited motion, and severely limited
or distorted motion, respectively.

\subsection{Judging Procedure}

For each generated video, we uniformly sample six temporally ordered
frames from the middle portion of the clip. The frames and two reference
images are resized so that their longest side does not exceed 512 pixels
and are JPEG-encoded. A single multimodal prompt contains the shared
five-axis rubric, the sampled frames, the product reference, and the human
reference. Method names are never included in the prompt.

The judge returns a JSON object containing five integer scores and a
one-sentence justification. All axes are evaluated in one call but are
explicitly instructed to be scored independently. Invalid responses are
retried up to five times. Parsed scores are clamped to $[0,5]$, and a
missing or unparseable score is assigned 0.

\subsection{Aggregation and Evaluation Controls}

For method $m$ and evaluation axis $a$, the normalized score is
\begin{equation}
S_{m,a}=\frac{1}{5N}\sum_{i=1}^{N}s_{m,a}^{(i)},
\end{equation}
where $s_{m,a}^{(i)}\in[0,5]$ is the judge score for case $i$ and
$N=150$ within each evaluation. The main comparison and ablation study
use separate held-out case sets. Dividing by 5 maps the reported results to $[0,1]$ while
preserving the ordering between methods. The 5D Avg. in the main
comparison is the unweighted arithmetic mean of the five normalized
axis scores. A missing video receives 0 on
all axes and remains in the average. For single-reference inference,
reference-dependent axes are marked as inapplicable and excluded from
the corresponding average.

All methods are evaluated using the same case set, frame-sampling
procedure, input resolution, prompt template, scoring rubric, and judge
model. The judge is blind to method identity, and the axes are scored
independently to reduce cross-axis influence.

\section{Dataset Construction Details}
\label{app:dataset}

\subsection{Asset Sources}
We construct the dataset using both generated and real-world product and presenter assets. For products, Qwen3.6-27B prompts GPT-Image-2 to synthesize 1,500 images, while real product assets are collected from e-commerce platforms and Omnitryon. For presenters, GPT-Image-2 generates 1,500 empty-hand portraits, which are supplemented with filtered real-world portraits.

\subsection{Asset Filtering}
Product images are processed using BiRefNet for foreground extraction and a VLM-based filter covering eight quality dimensions. Their masks and boundaries are subsequently corrected manually. Human images undergo pHash deduplication, YOLOv8-based single-subject filtering, SAM-based background assessment, and aesthetic filtering using a SCUT-FBP5500-finetuned ViT.

\subsection{Video Synthesis and Quality Control}
For each product--presenter pair, Qwen3.6-27B generates an e-commerce prompt describing spatial layout and temporal dynamics. GPT-Image-2 synthesizes the initial frame, and Wan2.7 generates the corresponding video. GPT-5.5 then filters candidate videos according to visual quality, temporal consistency, and prompt compliance. Product and face masks are precomputed for region-weighted training.

\subsection{Dataset Composition}
The final dataset contains 10,000 videos: 3,000 synthetic samples, 800 Omnitryon products paired with synthetic presenters, 2,000 Omnitryon products paired with real presenters, 3,000 real products paired with real presenters, and 1,200 real products paired with synthetic presenters.

\subsection{Source Use and Privacy}
Real-world assets are stored using anonymous internal identifiers without associated account names or unrelated metadata. Only visual content and derived masks required for model training are retained. The training and evaluation asset pools are strictly separated at both the product-instance and presenter-identity levels.

\end{document}